\documentclass[11pt,a4paper,fleqn]{article}
\usepackage[hyperref]{emnlp-ijcnlp-2019}
\usepackage{times}
\usepackage{latexsym}
\usepackage{url}
\usepackage{graphicx} 
\usepackage{latexsym}
\usepackage{multirow}
\usepackage{graphicx}
\usepackage{color}

\aclfinalcopy 

\title{GECOR: An End-to-End Generative Ellipsis and Co-reference Resolution Model for Task-Oriented Dialogue}

\author{Jun Quan\footnotemark[3]  \thanks{\ \  Work performed during an internship at Lenovo Research AI Lab.} ,\  Deyi Xiong\footnotemark[3]  \thanks{\ \  Corresponding author} ,\  Bonnie Webber\footnotemark[4] \and Changjian Hu\footnotemark[5] \\
  \footnotemark[3]\ \  School of Computer Science and Technology, Soochow University, Suzhou, China \\
  \footnotemark[4]\ \  University of Edinburgh, Edinburgh, UK \\
  \footnotemark[5]\ \  Lenovo Research AI Lab, Beijing, China \\
  {\tt terryqj0107@gmail.com},\quad {\tt dyxiong@suda.edu.cn},\\
  {\tt bonnie.webber@ed.ac.uk},\quad {\tt hucj1@lenovo.com} \\}

\date{}

\begin{document}
\maketitle
\begin{abstract}
  Ellipsis and co-reference are common and ubiquitous especially in multi-turn dialogues. In this paper, we treat the resolution of ellipsis and co-reference in dialogue as a problem of generating omitted or referred expressions from the dialogue context. We therefore propose a unified end-to-end Generative Ellipsis and CO-reference Resolution model (GECOR) in the context of dialogue. The model can generate a new pragmatically complete user utterance by alternating the generation and copy mode for each user utterance. A multi-task learning framework is further proposed to integrate the GECOR into an end-to-end task-oriented dialogue. In order to train both the GECOR and the multi-task learning framework, we manually construct a new dataset on the basis of the public dataset CamRest676 with both ellipsis and co-reference annotation. On this dataset, intrinsic evaluations on the resolution of ellipsis and co-reference show that the GECOR model significantly outperforms the sequence-to-sequence (seq2seq) baseline model in terms of EM, BLEU and F$_1$ while extrinsic evaluations on the downstream dialogue task demonstrate that our multi-task learning framework with GECOR achieves a higher success rate of task completion than TSCP, a state-of-the-art end-to-end task-oriented dialogue model \cite{lei2018sequicity}.
\end{abstract}

\section{Introduction}

Due to the rhetorical principle of saving words and avoiding repetitions, ellipsis and co-reference occur frequently in multi-turn dialogues leaving utterances paragmatically incomplete if they are separate from context. Humans can easily understand utterances with anaphorically referenced or absent  information (e.g., $Q_2$ and $Q_3$ in Table \ref{An example of recovery}) based on the dialogue context while dialogue systems often fail to understand such utterances correctly, which may result in false or incoherent responses.

If user utterances can be automatically supplemented with information that is left out or substituted by anaphora according to the dialogue context as humans do  (e.g., \emph{Q$_2$: I want cheap Italian restaurants.  Q$_3$: Yes, I would like the phone number please.}), dialogue models may understand user requests correctly and would not generate wrong responses caused by ellipsis and co-reference phenomena. Especially in task-oriented dialogue systems, explicitly providing such information to the models can effectively improve the success rate of task completion.

In order to achieve this goal, we propose an end-to-end generative ellipsis and co-reference resolution model (GECOR) for task-oriented dialogue in this paper. The essential idea behind GECOR is that we treat the resolution of ellipsis and co-reference in user utterances as a generation task: transforming a user utterance with ellipsis or anaphora into a new utterance where the left-out or referred expressions are automatically generated from the dialogue context. We refer to the new utterance as the complete version of the original utterance. We use an end-to-end sequence-to-sequence model with two encoders for this transformation task, where one encoder reads the user utterance and the other the dialogue context and the decoder generates the complete utterance. Since most omitted expressions or antecedents can be found in the dialogue context, we resort to the attention and copy mechanism to detect such fragments in previous context and copy them into the generated complete utterance.

We then incorporate GECOR into an end-to-end task-oriented dialogue system in a multi-task learning framework. The entire model contains two encoders (one for user utterance and the other for the dialogue context) and three decoders: one decoder for predicting dialogue states, the second decoder for generating complete user utterances and the third decoder for generating system responses. The three decoders are jointly trained.

In order to train GECOR with the task-oriented dialogue model, we manually annotate the public task-oriented dialogue dataset CamRest676 with omitted expressions and substitute anaphora in the dataset with corresponding antecedents. The new dataset can be used either to train a stand-alone ellipsis or co-reference resolution model or to jointly train a task-oriented dialogue model equipped with the ellipsis / co-reference resolution model.

We conduct a series of experiments and analyses, demonstrating that the proposed method can significantly outperform a strong baseline model. Our contributions are threefold: 
\begin{itemize}
\item We propose an end-to-end generative resolution model that attempts to solve the ellipsis and co-reference reolution in a single unified framework, significantly different from previous end-to-end co-reference resolution network with two phases of detection and candidate ranking.  
\item To the best of our knowledge, this is the first attempt to combine the task of ellipsis and co-reference resolution with the multi-turn task-oriented dialogue. The success rate of task completion is significantly improved with the assistance of the ellipsis and co-reference resolution. 
\item We construct a new dataset based on CamRest676 for ellipsis and co-reference resolution in the context of task-oriented dialogue.\footnote{The new dataset and the code of our proposed system are available at \url{https://multinlp.github.io/GECOR/}}
\end{itemize}

\section{Related Work}
\textbf{Ellipsis recovery:} The earliest work on ellipsis as far as we know is the PUNDIT system \cite{Palmer:1986:RII:1077146.1077160} which discusses the communication between the syntactic, semantic and pragmatic modules that is necessary for making implicit linguistic information explicit. \citet{dalrymple1991ellipsis} and \citet{shieber1996interactions} establish a set of linguistic theories in the ellipsis recovery of English verb phrases. \citet{nielsen2003corpus} first proposes an end-to-end computable system to perform English verb phrase ellipsis recovery on the original input text. \citet{liu2016exploring} propose to decompose the resolution of the verb phrase ellipsis into three sub-tasks: target detection, antecedent head resolution, and antecedent boundary detection. 

\textbf{Co-reference resolution:} Co-reference resolution is mainly concerned with two sub-tasks, referring expressions (i.e., mentions) detection, and entity candidate ranking. \citet{uryupina2013multilingual} propose a rule-based approach for co-reference detection which employs parse tree features with an SVM model. \citet{peng2015joint} improve the performance of mention detection by applying a binary classififier on their feature set. In recent years, applying deep neural networks to the co-reference resolution has gained great success. \citet{clark2016deep} apply reinforcement learning on mention-ranking co-reference resolution. \citet{lee2017end} introduce the first end-to-end co-reference resolution model. \citet{lee2018higher} present a high-order co-reference resolution model with coarse-to-fine inference.

\textbf{Ellipsis and co-reference resolution in QA and Dialogue:} The methods mentioned above do not generalize well to dialogues because they normally require a large amount of well-annotated contextual data with syntactic norms and candidate antecedents. In recent years, a few studies try to solve ellipsis / co-reference resolution tailored for dialogue or QA tasks. \citet{kumar2016non} train a semantic sequence model to learn semantic patterns and a syntactic sequence model to learn linguistic patterns to tackle with the non-sentential (incomplete) questions in a question answering system. \citet{zheng2018Sequence} builds a seq2seq neural network model for short texts to identify and recover ellipsis. However, these methods are still limited to short texts or one-shot dialogues. Our work is the first attempt to provide both solution and dataset for ellipsis and co-reference resolution in multi-turn dialogues.

\textbf{End-to-end task-oriented dialogue:} Task-oriented dialogue systems have evolved from traditional modularized pipeline architectures \cite{rudnicky1999creating,zue2000juplter,zue2000conversational} to recent end-to-end neural frameworks \cite{eric2017copy,eric2017key,lei2018sequicity,jin2018explicit}. Our work is an innovative combination of ellipsis and co-reference resolution and the end-to-end task-oriented dialogue.

\begin{table}[]
\centering
\resizebox{\textwidth}{!}{%
\scalebox{0.01}{
\begin{tabular}{ll}
\hline
\textbf{Turn} & \textbf{Dialogue} \\ \hline
Q$_1$ & I would like an Italian restaurant. \\
A$_1$ & What price range do you have in mind? \\
Q$_2$ & I want cheap ones. \\
\begin{tabular}[c]{@{}l@{}}
     A$_2$  \\ \\ \\
\end{tabular} & \begin{tabular}[c]{@{}l@{}}Pizza Hut Cherry Hinton serves Italian \\ food in the south part of town. Would \\ you like their phone number?\end{tabular} \\
Q$_3$ & Yes, please. \\ \hline
\multicolumn{2}{l}{\textbf{User utterances after resolution}} \\ \hline
Q$_2$ & I want cheap Italian restaurants. \\
Q$_3$ & Yes, I would like the phone number please. \\ \hline
\end{tabular}%
}
}
\caption{Examples of ellipsis and co-reference resolution}
\label{An example of recovery} 
\end{table}

\section{The GECOR Model}
\label{recovery}
In this section, we reformulate the ellipsis and co-reference resolution task in the context of multi-turn dialogue and detail the proposed GECOR model.

\begin{figure*}[t] 
\centering 
\includegraphics[scale=0.63]{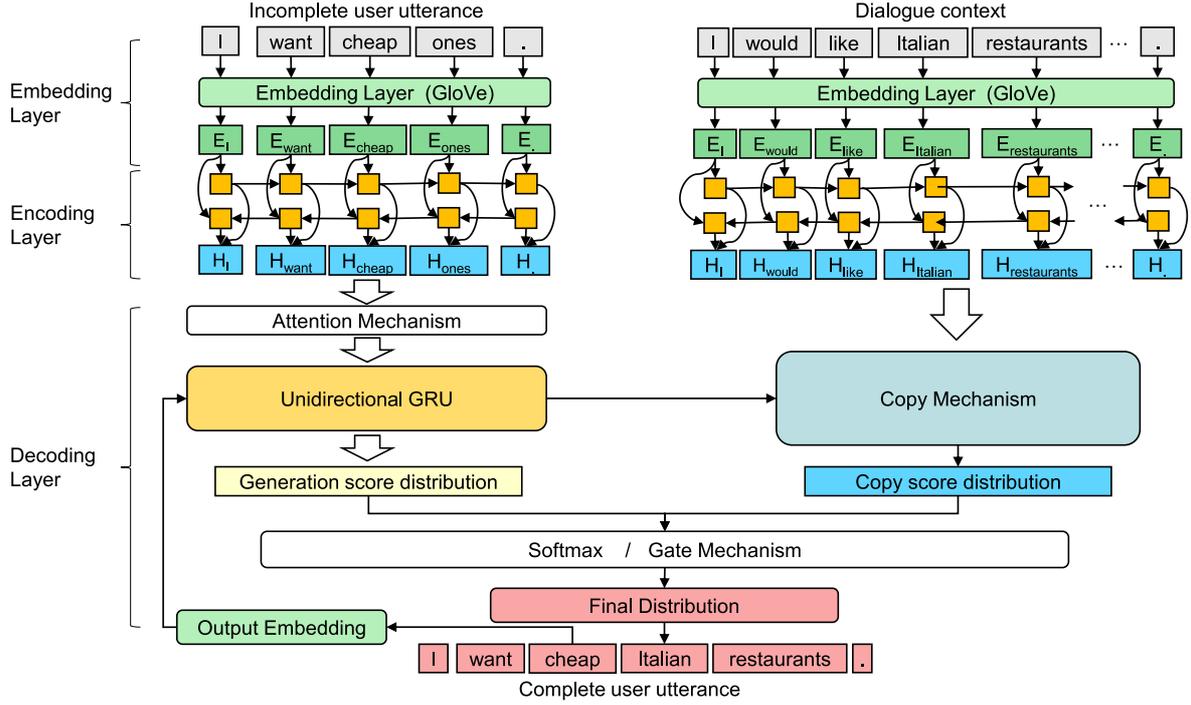}
\caption{The end-to-end generative model for ellipsis and co-reference resolution (GECOR).}
\label{recovery model} 
\end{figure*}

\subsection{Ellipsis and Co-Reference Resolution Reformulation}
Our task is to reconstruct a pragmatically complete utterance from a user utterance where the ellipsis and/or co-reference phenomena are present according to the dialogue context. Table \ref{An example of recovery} provides examples of reconstructed utterances in which the omitted information is recovered or  the anaphor is substituted with referred expressions.

We attempt to solve the resolution of ellipsis and co-reference in a unified framework because in essence both ellipsis and co-reference can be understood from contextual clues. We consider these two problems in multi-turn dialogue and reformulate the resolution of them as a generation problem: generating the omitted or referred expressions. In this way, the modeling of ellipsis and co-reference is in line with response generation in dialogue modeling.

Unlike previous methods that combine detection and ranking models, our
generation-based formulation is not constrained by the syntactic forms of ellipsis or co-reference in sentences. They can be either words (e.g., \emph{noun, verb}) or phrases or even clauses. Furthermore, the formulation does not need to provide a set of candidate antecedents to be resolved. Previous studies usually need to traverse the text when there are multiple ellipsis or anaphora to be resolved, which leads to a high computational complexity.

In this reformulation, we assume that the dialogue context is composed of all utterances from the beginning of the dialogue to the current user utterance. Both the context and the user utterance in question are input to the GECOR model to generate the complete version of the user utterance.

\subsection{Model Structure}
The GECOR model is shown in Figure \ref{recovery model}. The model essentially contains an embedding module, a user utterance encoder, a dialogue context encoder and a decoder with either copy \cite{gu2016incorporating} or gated copy mechanism (modified from \citet{see-etal-2017-get}). Both the generation probability over the entire vocabulary and the copy probability over all words from the dialogue context are taken into account for predicting the complete user utterance.

\textbf{Embedding Layer}  In GECOR, we first tokenize the input user utterance and the dialogue context. We then use GloVe \cite{pennington2014glove} (the pre-trained 50-dimensional word vectors) in the embedding layer to obtain word embeddings. Let \textbf{U} = \{\emph{u$_1$, ..., u$_m$}\}, \textbf{C} = \{\emph{c$_1$, ..., c$_n$}\} be representations of the tokenized utterance and context sequence. 

\textbf{Utterance and Context Encoder} We use a single-layer bidirectional GRU to construct both encoders. The forward and backward hidden states over the input embeddings from the embedding layer are concatenated to form the hidden states of the two encoders. 

\textbf{Decoder} The decoder is a single-layer unidirectional GRU. In the decoder, the attention distribution \emph{$a^t$} is calculated as in \citet{Bahdanau2015NeuralMT}:
\begin{equation}
e_i^t = v^T tanh(W_hh_i +W_ss_{t-1} +b_{attn})
\end{equation}
\begin{equation}
a^t = softmax(e^t)
\end{equation}
where \emph{$v$}, \emph{$W_h$}, \emph{$W_s$} and \emph{$b_{attn}$} are learnable parameters, $h_i$ is the hidden state for word $u_i$ from the sequence produced by the utterance encoder. The attention distribution \emph{$a^t$} is used to produce a weighted sum of the encoder hidden states, known as the context vector \emph{$h_t^*$}:
\begin{equation}
h_t^* = \sum_{i}a_i^th_i
\end{equation}
It is fed into the single-layer unidirectional GRU together with the previous decoder state \emph{$s_t$} and the word embedding $y_{t-1}$ of the previously generated word to obtain the decoder state $s_t$. The updated \emph{$s_{t-1}$} is then concatenated with the context vector \emph{$h_t^*$} to produce the generation probability distribution over the vocabulary \textbf{V} as follows:

\begin{equation}
    P^g(y_t)=\frac{1}{Z}e^{\psi_g(y_t)}, \quad    y_t\in \textbf{V}
\end{equation}

\begin{equation}
    \psi_g(y_t=v_i)=\textbf{v$_i$}^T (W_g^h h_t^* + W_g^s s_t+b_g)
\end{equation}

\begin{equation}
    s_t=GRU([y_{t-1};h_t^*],s_{t-1})
\end{equation}
where \emph{W$_g^h$}, \emph{W$_g^s$} and \emph{b$_g$} are learnable parameters and \textbf{v$_i$} is the one-hot indicator vector for word  $v_i \in \textbf{V}$.  $\psi_g$ is the score function for the generation-mode and $Z$ is the normalization term shared by the generation-mode and copy-mode.

\textbf{Copy Network} The copy network \cite{gu2016incorporating} is used to calculate the probabilities for words copied from the dialogue context. These words are parts of the omitted or referred expressions to be predicted. We build the copy network on the top of the context encoder. The probability for copying each word from the dialogue context is computed as follows:

\begin{equation}
    P^c(y_t)=\frac{1}{Z}\sum_{i:c_i=y_t}^{|\textbf{C}|} e^{\psi_c(c_i)},\quad    y_t\in \textbf{C}
\end{equation}

\begin{equation}
    \psi_c(y_t=c_i)=\sigma (W_{c}h_i^c+b_c)s_t
\end{equation}
where \emph{W$_c$} and \emph{b$_c$} are learnable parameters, $h_i^{c}$ is the output for word $c_i$ from the context encoder, and \emph{$\sigma$} is a non-linear activation function. $\psi_c$ is the score function for the copy-mode and $Z$ is the normalization term shared by equation (4) and (7).

Both probabilities from the two modes contribute to the final probability distribution over the extended vocabulary (the vocabulary plus the words from the dialogue context) which is calculated as follows: 
\begin{equation}
    P(y_t)=P^g(y_t)+P^c(y_t), \   y_t\in \textbf{V}\cup \textbf{C}
\end{equation}
which is used to predict the final output word.

\textbf{Gated Copy} An alternative to the copy network is the gated copy mechanism that use a gate to regulate the contributions of the generation and copy mode to the final prediction. The gate $p_{gen}$ is calculated as follows:
\begin{equation}
    p_{gen} = \sigma(W_{h} h_t^* + W_{s} s_t + W_{y} y_{t-1} + b_t)
\end{equation}
\begin{equation}
    P(y_t) = p_{gen}P^g(y_t) + (1-p_{gen})P^c(y_t)
\end{equation}
where $W_h$, $W_s$, $W_y$ and $b_t$ are learnable parameters and $\sigma$ is the sigmoid function.

\textbf{Training} The standard cross-entropy loss is adopted as the loss function to train the GECOR model.

\begin{figure*}[t] 
\centering 
\includegraphics[scale=0.62]{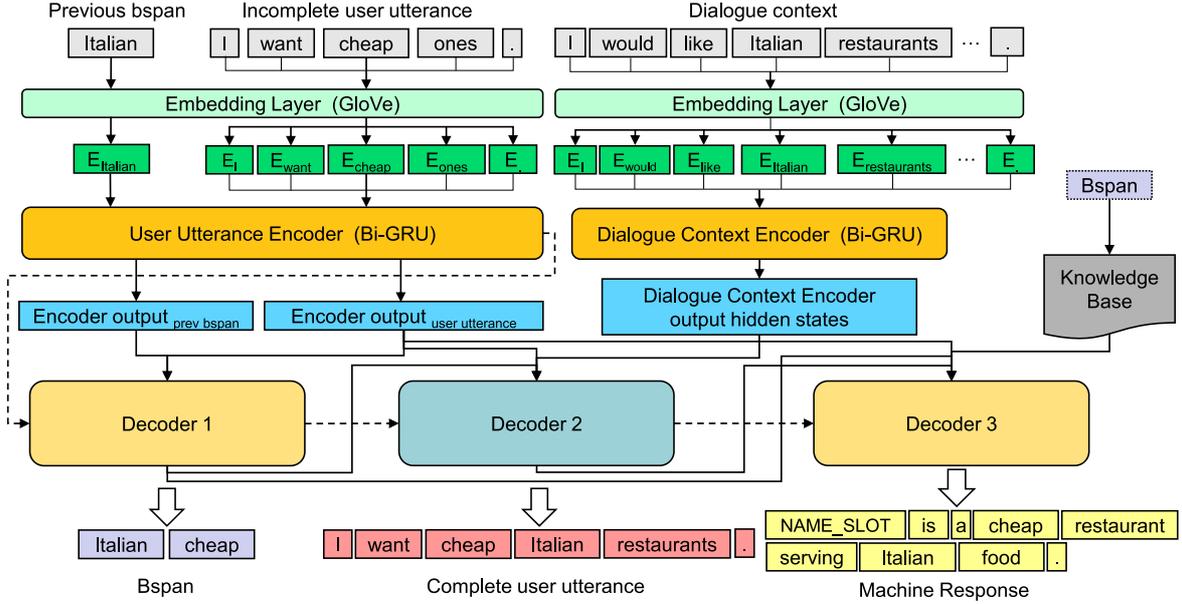}
\caption{The architecture of the end-to-end task-oriented dialogue enhanced with the GECOR model. Decoder 1: BSpan decoder. Decoder 2: completed user utterance decoder. Decoder 3: machine response decoder. }
\label{multi-task learning model} 
\end{figure*}

\section{Task-Oriented Dialogue with GECOR}

We integrate the proposed GECOR into an end-to-end task-oriented dialogue system TSCP proposed by \citet{lei2018sequicity} in a multi-task learning framework, which is shown in Figure \ref{multi-task learning model}. The GECOR-equipped TSCP model contains the embedding layer, the utterance and context encoders, and three decoders: decoder 1 for generating belief spans (BSpan) defined in \cite{lei2018sequicity} which are text spans for tracking dialogue states (e.g., \emph{$\langle inf \rangle  Italian, cheap \langle /inf \rangle ;\langle req \rangle  phone \langle /req \rangle $}), decoder 2 for complete user utterances and decoder 3 for machine responses. The embedding layer and encoders are the same as described in section \ref{recovery}.

\textbf{BSpan Decoder} Unlike \citet{lei2018sequicity}, we do not concatenate current user utterance with previously generated machine response. At each turn of dialogue, the user utterance and the previous BSpan (the dialogue states updated to the previous turn) are used as the inputs to the user utterance encoder. The outputs of this encoder are then fed into the BSpan decoder for predicting the new BSpan for the current turn and a cross-entropy loss \emph{$L_1$} is calculated. The user utterance encoder hidden states, the last hidden state and the output of the BSpan decoder are input into the other two decoders. 

\textbf{Complete User Utterance Decoder} The basic structure of this decoder is the same as the decoder described in the last section.We pass the last hidden state of the BSpan decoder to the initial state of this decoder. In addition to the inputs from the user utterance encoder and the dialogue context encoder, we also input the output of the BSpan decoder into this decoder. The generation probability \emph{P$_t^g$}, copy probability \emph{$P_t^{c1}$} for copying tokens in BSpan, and copy probability \emph{$P_t^{c2}$} for copying words in the dialogue context are calculated with a shared normalization term and combined for the final probability computation:
\begin{equation}
    P_t = P_t^g + P_t^{c1} + P_t^{c2}
\end{equation}
\emph{$P_t$} is then used to decode the words in the complete user utterance. For this decoder, the second cross-entropy loss \emph{$L_2$} is calculated. 

\textbf{Machine Response Decoder} Similar to the previous two decoders, the machine response decoder is also a single-layer unidirectional GRU, the initial state of which is set to the last hidden state of the complete user utterance decoder. In this decoder, we compute three context vectors for each decoder state \emph{$s_t$}. The first context vector \emph{$h_{t1}^*$} is calculated over the user utterance encoder hidden states while the other two context vectors \emph{$h_{t2}^*$}, \emph{$h_{t3}^*$} are calculated over the BSpan decoder hidden states and the complete user utterance decoder hidden states, respectively. The concatenation of \emph{$s_t$}, \emph{$h_{t1}^*$}, \emph{$h_{t2}^*$}, \emph{$h_{t3}^*$} and the Knowledge Base (KB) matching vector \emph{$K_t$} (a one-hot representation of the retrieval results in KB according to the constraints in the corresponding BSpan) is used to generate the output and update the decoder state. The generated output is then concatenated with the three context vectors to feed into a layer to produce the generation probability distribution over the vocabulary. Similar to the complete user utterance decoder, we also use the copy mechanism in the machine response decoder. The third cross-entropy loss \emph{$L_3$} is then calculated. 

\textbf{Training} The final loss for the multi-task learning framework is estimated as follows:
\begin{equation}
    L = L_1+L_2+L_3
\end{equation}
We learn parameters to minimize the final loss.

\section{Data Annotation for Ellipsis and Co-Reference Rosultion in Dialogue }
Since there are no publicly available labeled data for the resolution of ellipsis and co-reference in dialogue, we manually annotate such a new dataset based on the public dataset CamRest676 \cite{wen2016conditional,wen2016network} from the restaurant domain.

\begin{table}[]
\centering
\resizebox{\textwidth}{!}{%
\scalebox{0.009}{
\begin{tabular}{ll}
\hline
\textbf{Turn} & \textbf{Dialogue} \\ \hline
Q$_1$ & I would like a traditional food restaurant. \\
A$_1$ & What price range do you have in mind? \\
Q$_2$ & I don't care. \\ \hline
Q$_2$ (Complete) & I don't care about the price range. \\
Q$_2$ (Ellipsis) & I don't care. \\
Q$_2$ (Co-reference) & I don't care about it. \\ \hline
\end{tabular}%
}
}
\caption{An example of the ellipsis / co-reference annotation}
\label{annotation example}
\end{table}

\textbf{Annotation Specification} Annotation cases for user utterances can be summarized into the following three conventions:
\begin{itemize}
    \item As shown in Table \ref{annotation example}, if a user utterance contains an ellipsis or anaphor, we manually resolve the ambiguity of ellipsis or anaphor and supplement the user utterance with a correct expression by checking the dialogue context. In doing so, we create a pragmatically \emph{complete} version for the utterance. If the utterance only contains an ellipsis and the ellipsis can be replaced with an anaphor, we create a \emph{co-reference} version for it. Similarly, if the utterance only contains an anaphor and the anaphor can be omitted, we create an \emph{ellipsis} version for the utterance. 
    \item If the user utterance itself is pragmatically complete, without any ellipsis or anaphora, we create an anaphor and ellipsis version for it if such a creation is appropriate. 
    \item If the utterance itself is complete and it is not suitable to create an ellipsis or anaphor version, we just do nothing.
\end{itemize}
With the annotation convention described above, for each user utterance in the dataset, we can label it as $l\in$ \{\emph{ellipsis}, \emph{co-reference}, \emph{complete}\} or create two other versions for it if appropriate. Please notice that these labels are used only for dataset statistics or for designing experiments, not for training our models.

\textbf{Dataset statistics} The CamRest676 dataset contains 676 dialogues, with 2,744 user utterances. After annotation, 1,174 \emph{ellipsis} versions and 1,209 \emph{co-reference} versions are created from the 2,744 user utterances. 1,331 incomplete utterances are created that they are an either \emph{ellipsis} or \emph{co-reference} version. 1,413 of the 2,744 user utterances are complete and not amenable to change. No new versions are created from these 1,413 utterances.

\textbf{Dataset Split for Experiments} We split the new dataset into a training set (accounting for 80\%) and validation set (accounting for 20\%) which can be used for the stand-alone ellipsis/co-reference resolution task and the multi-task learning of both the ellipsis/co-reference resolution and end-to-end task-oriented dialogue.

\begin{table*}[ht]
\centering
\resizebox{\textwidth}{!}{%
\scalebox{0.19}{
\begin{tabular}{|l|l|l|l|l|l|l|l|l|l|}
\hline
\multirow{2}{*}{\textbf{Data}} & \multirow{2}{*}{\textbf{Model}} & \multicolumn{8}{c|}{\textbf{Resolution Task}} \\ \cline{3-10} 
 &  & \textbf{EM(\%)} & \textbf{EM 1(\%)} & \textbf{EM 2(\%)} & \textbf{BLEU(\%)} & \textbf{F$_1$(\%)} & \textbf{Prec.(\%)} & \textbf{Rec.(\%)} & \textbf{Reso.F$_1$(\%)} \\ \hline
\multirow{3}{*}{Ellipsis} & Baseline & 49.99 & 68.88 & 27.31 & 73.26 & 90.89 & 92.14 & 89.67 & 44.47 \\ \cline{2-10} 
 & GECOR 1 & 67.56 & 92.07 & 37.18 & 83.69 & 96.25 & 98.28 & 94.30 & 70.48 \\ \cline{2-10} 
 & GECOR 2 & 67.75 & 91.38 & 38.46 & 82.94 & 96.58 & 98.48 & 94.76 & 70.85 \\ \hline
\multirow{3}{*}{Co-reference} & Baseline & 55.64 & 76.03 & 33.60 & 78.12 & 92.58 & 93.28 & 91.89 & 44.24 \\ \cline{2-10} 
 & GECOR 1 & 71.35 & 91.67 & 47.68 & 85.89 & 96.49 & 98.19 & 94.86 & 64.93 \\ \cline{2-10} 
 & GECOR 2 & 71.18 & 93.80 & 44.92 & 85.93 & 97.09 & 98.46 & 95.76 & 71.26 \\ \hline
\multirow{3}{*}{Mixed} & Baseline & 50.38 & 70.89 & 28.57 & 74.11 & 90.93 & 91.72 & 90.15 & 44.10 \\ \cline{2-10} 
 & GECOR 1 & 68.52 & 92.03 & 42.04 & 83.91 & 95.88 & 98.12 & 93.74 & 66.06 \\ \cline{2-10} 
 & GECOR 2 & 66.22 & 91.64 & 37.45 & 82.98 & 96.47 & 98.41 & 94.60 & 66.16 \\ \hline
\end{tabular}%
}
}
\caption{Results of the resolution task on the dataset. GECOR 1/2: the GECOR model with the copy/gated copy mechanism. EM 1 and EM 2 respectively indicate the situation that the input utterance is complete or incomplete while EM is the comprehensive evaluation of the two situations. Reso.F$_1$: Resolution\_F$_1$}
\label{results of recovery model} 
\end{table*}

\section{Experiments}
In this section we conducted experiments on the new dataset to examine the generative ellipsis/co-reference resolution model and its integration into the end-to-end task-oriented dialogue.

\subsection{Evaluation Metrics}
As far as we know, there is no end-to-end generative ellipsis and co-reference resolution model applied to multi-turn dialogues. Therefore there are no off-the-shelf metrics tailored to this evaluation. Since we deal with two tasks: the task of ellipsis/co-reference resolution (resolution task for short) and the task-oriented dialogue with integrated ellipsis/co-reference resolution (hereafter dialogue task), we use two sets of evaluation metrics. For the resolution task, we use the exact match rate (EM) that measures whether the generated utterances exatly match the gold utterances.

BLEU \cite{papineni2002bleu} and F$_1$ score (a balance between word-level precision and recall) are also used for the resolution task to evaluate the quality of generated utterances at the n-gram and word level. We use the success F$_1$ which is defined as the F$_1$ score of requested slots correctly answered in dialogues to evaluate task completion rate for the dialogue task, similar to  \citet{lei2018sequicity}.

\subsection{Parameter Settings}
For all our models, both the size of hidden states and word embeddings were set to 50. The vocabulary size $|$\textbf{V}$|$ was set to 800 and the batch size was set to 32. We trained our models via the Adam optimizer \cite{kingma2015adam}, with a learning rate of 0.003 and a decay parameter of 0.5. Early stopping and dropout were used to prevent overfitting, and the dropout rate was set to 0.5.

\subsection{Baselines and Comparisons}
For the resolution task, we compared our GECOR model with the baseline model proposed by \citet{zheng2018Sequence} which is a seq2seq neural network model that identifies and recovers ellipsis for short texts.

For the dialogue task, we compared our multi-task learning framework with the baseline model TSCP proposed by \citet{lei2018sequicity} which is a seq2seq model enhanced with reinforcement learning. We ran the source code\footnote{https://github.com/WING-NUS/sequicity} on our dataset to get the baseline results for comparison. 

For the resolution task, we also performed a comparison study to examine the impacts of the gate mechanism incorporated into the copy network on the GECOR model and on the multi-task learning dialogue model.

\begin{table*}[t]
\centering
\resizebox{\textwidth}{!}{%
\begin{tabular}{|l|l|l|l|l|l|l|l|l|l|}
\hline
\multirow{2}{*}{\textbf{Data}} & \multirow{2}{*}{\textbf{Model}} & \multicolumn{5}{c|}{\textbf{Resolution Task}} & \multicolumn{3}{c|}{\textbf{Dialogue Task}} \\ \cline{3-10} 
 &  & \textbf{EM(\%)} & \textbf{BLEU(\%)} & \textbf{F1(\%)} & \textbf{Prec.(\%)} & \textbf{Rec.(\%)} & \textbf{Succ.F1(\%)} & \textbf{Prec.(\%)} & \textbf{Rec.(\%)} \\ \hline
Complete & TSCP & - & - & - & - & - & 86.30 & 89.60 & 83.23 \\ \hline
\multirow{2}{*}{Ellipsis} & TSCP & - & - & - & - & - & 84.56 & 87.25 & 82.02 \\ \cline{2-10} 
 & Our Model & 60.83 & 78.89 & 95.64 & 97.79 & 93.58 & \textbf{85.33} & 88.69 & 82.21 \\ \hline
\multirow{2}{*}{Co-reference} & TSCP & - & - & - & - & - & 82.17 & 88.91 & 76.38 \\ \cline{2-10} 
 & Our Model & 68.56 & 83.98 & 96.61 & 98.09 & 95.18 & \textbf{86.00} & 90.46 & 81.95 \\ \hline
\multirow{2}{*}{Mixed} & TSCP & - & - & - & - & - & 83.25 & 86.91 & 79.89 \\ \cline{2-10} 
 & Our Model & 66.47 & 83.63 & 96.26 & 98.16 & 94.44 & \textbf{85.97} & 87.98 & 84.05 \\ \hline
\end{tabular}%
}
\caption{Results of the multi-task learning model. This table is split into two parts: performance of resolution for the integrated GECOR on the left side and performance of dialogue task on the right side.}
\label{results of our multi-task learning model} 
\end{table*}

\subsection{The GECOR Model}
Our generative resolution model was trained on three types of data: the ellipsis data where only ellipsis version utterances from the annotated dataset were used, the co-reference data where only co-reference version utterances from the annotated dataset were used, and the mixed data where we randomly selected a version for each user utterance from \{\emph{ellipsis}, \emph{co-reference}, \emph{complete}\}. In the mixed data, 633 turns are with ellipsis user utterances, 698 turns are with co-reference user utterances, and the rest are with complete user utterances. The experimental results of the GECOR and baseline model \cite{zheng2018Sequence} on the three different datasets are shown in Table \ref{results of recovery model}. 

\textbf{Overall results} From the third column of the table, we find that the GECOR model with the copy mechanism (GECOR 1) improves the exact match rate (EM) by more than 17 points on the ellipsis version data, more than 15 points on the co-reference data, and more than 18 points on the mixed data. We further define a metric we term as \textbf{Resolution\_F$_1$} that is an F$_1$ score calculated by comparing machine-generated words with ground truth words for only the ellipsis / co-reference part of user utterances. The GECOR model achieves consistent and significant improvements over the baseline in terms of BLEU, F$_1$ and Resolution\_F$_1$ in addition to the EM metric . The major difference between the GECOR and the baseline is that the former tries to copy words from the dialogue context. The improvements, especially the improvements on the ellipsis resolution (higher than those on the co-reference resolution) indicate that the copy mechanism is crucial for the recovery of ellipsis and co-reference. 

\textbf{Effect of the two copy mechanisms } Comparing the GECOR 1 against the GECOR 2 (with the gated copy mechanism), we can find that the gating between copy and generation is helpful in terms of the word-level quality (F$_1$ and Resolution\_F$_1$ score) but not in terms of the fragment or sequence-based metrics (i.e., BLEU and EM). Therefore, we only integrate the GECOR model with the copy mechanism into the dialogue system. 

\textbf{Incomplete vs. complete utterances } In multi-turn dialogues, user utterances may be incomplete or complete. A robust resolution model needs to be able to accurately identify whether the input utterance is complete or not. The model needs to keep it unchanged when it is complete and to predict the corresponding complete version when it is incomplete. For these cases, we tested our models and made statistical analysis on the three versions of data as shown in column 3, 4 and 5 of Table \ref{results of recovery model} (EM, EM 1, EM 2). We can find that the GECOR model beats the baseline model in all respects. However, the GECOR model needs further improvement when the input utterances are incomplete, compared with its good performance on complete utterances.

\textbf{Analysis on GECOR results for complete utterances } We then analyzed the experimental results of the GECOR 1 on the mixed data in detail. When the input user utterances are complete, the GECOR model can amazingly generate 92.03\% utterances that exactly match the input utterances. Only 7.97\% do not match perfectly. Most unmatched cases, as we found, are with: \textbf{(1) missed words} (e.g., User: \emph{Can I get a Korean restaurant in the town centre?} GECOR: \emph{Can I get a Korean restaurant in the town?}) \textbf{(2) Repetition} (e.g., User: \emph{OK, thank you. That is all for today then.} GECOR: \emph{OK, thank you. That is all for today for today then.})

\textbf{Analysis on GECOR results for incomplete utterances } For incomplete input user utterances, GECOR can generate 42.04\% exactly matched cases. Among the 57.96\% cases that do not exactly match ground truth utterances, only 6.3\% are not complete, which still contains unresolved ellipsis or co-reference, while 93.7\% of these cases are complete with GECOR-generated words that do not match ground truth words. An in-depth analysis on these show that they can be clustered into 4 classes. \textbf{(1) Paraphrases.} We found that the majority of the unmatched complete utterances generated by GECOR are actually paraphrases to the ground truth complete utterances (e.g., User: \emph{Any will be fine.} GECOR: \emph{Any food type will be fine.} Reference: \emph{Any type of restaurant will be fine.}).  This is also confirmed by the high scores of the word-level evaluation metrics in Table \ref{results of recovery model}. \textbf{(2) Partial resolution.} When a pronoun refers to more than one items, GECOR sometimes generate a partial resolution for the pronoun (e.g., User: \emph{I do not care about them.} GECOR: \emph{I do not care about the price range.} Reference: \emph{I do not care about the price range or location.}). \textbf{(3) Minor errors.} In a few cases, the resolution part is correct while there are some errors elsewhere. (e.g., User: \emph{How about Chinese food?} Prediction: \emph{How about international food on the south side of town?} Reference: \emph{How about Chinese food on the south side of town?}) \textbf{(4) Repetition.} Some cases contain repeatedly generated words.

We think that although not exactly matched, paraphrased complete utterances generated by GECOR are acceptable. These utterances are helpful for the downstream dialogue task. For other errors, such as partial resolution or repetition, it may be necessary to enhance the attention or copy mechanism further in GECOR.

\subsection{The Multi-Task Learning Model}
We further conducted experiments to extrinsically evaluate the GECOR model in task-oriented dialogue with the success F$_1$ metric. This is also to evaluate our multi-task learning framework in integrating the GECOR model into the end-to-end dialogue model. In addition to training the baseline TSCP model on the ellipsis, co-reference and mixed dataset, we also trained it on the dataset with only complete user utterances. This is to examine the ability of the baseline model in using correct contextual information presented in user utterances. The experimental results are shown in Table \ref{results of our multi-task learning model}.

\textbf{Overall results } In comparison to the baseline, we can see that our model improves the success F$_1$ score by nearly 4 points on the co-reference dataset, which is close to the score obtained by the baseline trained with the complete user utterances. On the mixed and ellipsis dataset, our model also achieves 2.7 points and 0.8 points of success F$_1$ score improvements, respectively. 

\textbf{Resolution performance of the integrated GECOR } We also provide the performance of the integrated GECOR on the resolution task in Table \ref{results of our multi-task learning model}. The performance is slightly lower than when the GECOR is trained independently as a stand-alone system.  This suggests that the GECOR is able to perform well when integrated into a dialogue system. The overall results demonstrate that the proposed multi-task learning framework for the end-to-end dialogue is able to improve the task completion rate by incorporating an auxiliary ellipsis/co-reference resolution task.

Since the BSpan decoder is also used in the baseline system to capture contextual information and track dialogue states, we believe that our multi-task learning model with the integrated GECOR will play a more important role in end-to-end dialgoue models that do not use state tracking modules, e.g., neural open-domain conversation models \cite{Vinyals2015A,li2016deep}.

\section{Conclusion and Future Work}
In this paper, we have extensively investigated the ellipsis and co-reference resolution in the context of multi-turn task-oriented dialogues. We have presented the GECOR, a unified end-to-end generative model for both ellipsis and co-reference resolution in multi-turn dialogues. A multi-task learning framework is further proposed to integrate the GECOR into the end-to-end task-oriented dialogue. In order to train and test the proposed model and framework, we manually created a new dataset with annotated ellipsis and co-reference information based on the publicly available CamRest676 dataset. Experiments on the resolution task show that the GECOR is able to significantly improve the performance in terms of the exact match rate, BLEU and word-level F$_1$ score. Experiments on the dialogue task demonstrate that the task completion rate of the task-oriented dialogue system is significantly improved with the aid of ellipsis and co-reference resolution.

Our work could be extended to end-to-end open-domain multi-turn dialogue. We will further improve our model by incorporating syntactic and location information. We would also like to adapt the proposed methods to document-level neural machine translation in the future.

\section*{Acknowledgments}
The present research was supported by the National Natural Science Foundation of China (Grant No.61861130364) and the Royal
Society (London) (NAF$\backslash$R1$\backslash$180122). We would like to thank the anonymous reviewers for their insightful comments.

\bibliography{emnlp-ijcnlp-2019}
\bibliographystyle{acl_natbib}

\end{document}